\documentclass[10pt,twocolumn]{article}

\usepackage[margin=0.78in]{geometry}
\usepackage{amsmath}
\usepackage{amssymb}
\usepackage[T1]{fontenc}
\usepackage{newtxtext}          

\usepackage{newtxmath}          
\usepackage[protrusion=true,expansion=true]{microtype}
\usepackage{graphicx}
\usepackage{booktabs}
\usepackage{array}
\usepackage{enumitem}
\usepackage{xcolor}
\usepackage{caption}
\usepackage{tikz}
\usepackage{pgfplots}
\pgfplotsset{compat=1.18}
\usepackage[colorlinks=true,allcolors=blue!55!black]{hyperref}
\usepackage{url}
\usepackage{placeins}   
\usepackage{titlesec}
\usepackage{balance}    
\usetikzlibrary{arrows.meta,positioning}

\titlespacing*{\section}{0pt}{1.5ex plus .3ex minus .2ex}{0.9ex plus .2ex}
\titlespacing*{\subsection}{0pt}{1.2ex plus .3ex minus .2ex}{0.6ex plus .2ex}
\titleformat{\section}{\normalfont\large\bfseries}{\thesection}{0.6em}{}
\titleformat{\subsection}{\normalfont\normalsize\bfseries}{\thesubsection}{0.6em}{}

\titleformat{\paragraph}[runin]{\normalfont\bfseries}{}{0pt}{}
\titlespacing{\paragraph}{\parindent}{0pt}{0.5em}

\newcommand{\code}[1]{\texttt{\small #1}}

\newcommand{\figorbox}[3]{%
  \IfFileExists{#1}{\includegraphics[width=#2]{#1}}%
  {\fbox{\parbox[c][2.8cm][c]{#2}{\centering\color{gray}\itshape #3\\(add image to paper/figures/)}}}}

\definecolor{mirrorblue}{RGB}{31,86,145}
\definecolor{mirrorgray}{RGB}{90,96,104}
\definecolor{boxbg}{RGB}{237,243,250}

\title{MIRROR}   
\hypersetup{pdftitle={MIRROR: Multimodal Intelligent Radiology Reasoning and Observation Reporter},
            pdfauthor={Vignesh Nagarajan}}

\begin{document}

\twocolumn[{%
\centering
\vspace{6pt}
{\Huge\bfseries MIRROR}\\[6pt]
{\large\bfseries Multimodal Intelligent Radiology Reasoning and Observation Reporter}\\[10pt]
{\itshape\normalsize A three-layer radiology pipeline whose language layer reads
structured evidence rather than pixels; three modality taxonomies are registered
and routed, one of which, chest X-ray, is trained and benchmarked here.}\\[18pt]

\normalsize
\begin{minipage}[t]{0.45\textwidth}
\centering
{\large\textbf{Vignesh Nagarajan}}\\[3pt]
{\small\itshape Corresponding Author}\\[4pt]
{\small Brown Foundation Scholar, Texas A\&M University}\\[2pt]
{\small Department of Computer Science \& Engineering}\\[2pt]
{\small\texttt{vigneshn26@tamu.edu}}
\end{minipage}%
\hspace{0.06\textwidth}%
\begin{minipage}[t]{0.45\textwidth}
\centering
{\large\textbf{Sriram Venkatapathy}}\\[3pt]
{\small\itshape Research Mentor}\\[4pt]
{\small Applied AI Researcher, Capital One}\\[2pt]
{\small PhD in Computer Science, IIT Hyderabad}\\[2pt]
{\small\texttt{sriram.venkatapathy@gmail.com}}
\end{minipage}\\[14pt]
{\small Global Indian Scientists \& Technocrats (GIST) 2026 Summer Research Internship}\\[20pt]

\begin{minipage}{0.88\textwidth}
\small
\textbf{Abstract.}
A radiologist reading a model's output faces two problems. The model returns a
number and no reason, and any system that turns that number into readable prose
can quietly add claims the model never made. \textbf{MIRROR} is a research
prototype built to separate those failures. It chains a multi-label classifier, a
Grad-CAM localizer that turns each positive finding into a named anatomical
region, and a report writer that receives the labels, probabilities, and regions
but never the image. Because the language layer cannot see pixels, it cannot
assert a finding the classifier did not make. We are precise about what that buys:
a MIRROR report's findings are auditable against the probability vector, while the
sentences framing them are ordinary generated text, and we show one stating a
cardiothoracic ratio the system never measured. One registry holds the taxonomy,
anatomy, and phrasing for chest X-ray, brain MRI, and head CT, so adding a
modality is a data change; all three are routed and tested, one is trained. On
ChestMNIST that classifier reaches macro
AUROC $0.729$ and ranks better than chance on all 14 labels, at $1.6$ to $6.8$
times the precision a random ranker would get. Yet at the default $0.5$ threshold
it emits no positive prediction at all for 11 of them, and its excellent-looking
Brier score of $0.045$ sits beside the $0.047$ earned by a predictor that ignores
the image. The discrimination is real; the decisions are not. Under the class
imbalance normal in radiology, aggregate metrics flatter models that do nothing,
and should be reported against that floor.
\end{minipage}
\vspace{10pt}
}]

\section{Introduction}

Deep networks now match specialists on narrow chest-radiograph reading
tasks~\cite{rajpurkar2017chexnet}, yet clinical adoption remains limited by a
trust gap: a bare probability tells a radiologist neither \emph{where} the model
is looking nor \emph{why} it concluded what it did, and high-stakes decisions
demand models that can be interrogated~\cite{rudin2019stop}. Saliency methods such
as Grad-CAM~\cite{selvaraju2017gradcam} expose \emph{where} a prediction comes
from, and language models can render structured evidence as prose, but the two are
rarely composed into one system whose outputs stay mutually consistent end to end.

Composing them introduces a second risk that the first does not have. A report
generator with access to the image can write fluent radiology prose asserting
findings no classifier ever produced, and such prose is difficult to audit
precisely because it reads well. \textbf{MIRROR} is built around a constraint
aimed at that risk: image $\rightarrow$ prediction $\rightarrow$ localization
$\rightarrow$ reasoning $\rightarrow$ report, with the language layer receiving
only the structured evidence produced upstream, never the raw image. The report
can therefore assert only findings the classifier predicted and the localizer
situated. Because only the finding taxonomy, the anatomical vocabulary, and a
little report phrasing differ between imaging modalities, those three things are
centralized in a single registry, and a study is routed to the correct classifier
head either explicitly or from the modality tag carried in its DICOM header.

\paragraph{Research question.} Can a radiology pipeline be built so that its
language layer is structurally incapable of asserting a finding the classifier did
not detect, and what does that constraint cost? We answer the first half
architecturally and the second empirically, on chest radiographs. Two things this
paper does \emph{not} answer should be stated at the outset. It does not measure
explanation \emph{quality}: our harness implements a pointing-game and IoU
protocol against the NIH lesion boxes, but running it needs a checkpoint trained
on the full-resolution release, so no box-level score is reported here and none
should be inferred. Nor does it measure whether any of this helps a human, which
would require a reader study with radiologists timed and scored on the same
studies with and without the explanation and report layers.

\paragraph{Contributions.}
\begin{enumerate}[leftmargin=1.3em,itemsep=1pt,topsep=2pt]
  \item The distinction between \emph{finding-level} grounding, which constraining
  the language layer to structured evidence does buy and which is checkable, and
  \emph{detail-level} grounding, which it does not buy and which we show failing
  in a real generated report.
  \item A registry-driven multimodality in which a modality's taxonomy, anatomical
  vocabulary, and report phrasing are data rather than code. Three taxonomies are
  registered, routed, and exercised by the test suite; one is benchmarked.
  \item An open-source implementation: three backbones, two explainability
  methods, an LLM report backend with a deterministic offline fallback, a real
  DICOM ingest, and two very different inference engines behind one response
  contract.
  \item A measured chest-radiograph result in which discrimination is real on
  every label while decisions are absent on almost all of them, and in which two
  aggregate metrics, Brier and AUPRC, only become interpretable once they are
  written against the no-skill floors their label prevalences imply.
\end{enumerate}

\section{Literature Review}
\label{sec:litreview}

\paragraph{Datasets and classification.} Large labeled chest-radiograph corpora
made supervised deep learning on this modality practical. The NIH ChestX-ray8/14
release~\cite{wang2017chestxray} provides $112{,}120$ frontal radiographs over 14
pathologies and remains a standard benchmark; CheXpert~\cite{irvin2019chexpert}
and MIMIC-CXR~\cite{johnson2019mimiccxr} extended scale, uncertainty labeling, and
paired free-text reports. CheXNet~\cite{rajpurkar2017chexnet} showed a 121-layer
DenseNet~\cite{huang2017densenet} reaching radiologist-level pneumonia detection.
EfficientNet~\cite{tan2019efficientnet} and Vision
Transformers~\cite{dosovitskiy2021vit} complete the standard toolkit; MIRROR keeps
all three interchangeable, with DenseNet-121 primary for comparability.

\paragraph{Visual explanation.} Class Activation
Mapping~\cite{zhou2016cam} localized the evidence behind a CNN prediction;
Grad-CAM~\cite{selvaraju2017gradcam} generalized it to arbitrary architectures via
gradients, and Score-CAM~\cite{wang2020scorecam} removed the gradient dependence
with perturbation-based weighting. Model-agnostic attribution methods, notably
LIME~\cite{ribeiro2016lime} and SHAP~\cite{lundberg2017shap}, explain individual
predictions without access to model internals. The pointing-game
protocol~\cite{zhang2016pointing} and intersection-over-union against annotated
lesions turn qualitative heatmaps into quantitative localization scores; our
harness implements both, though we do not run them here.

\paragraph{Critiques of post-hoc explanation.} Adebayo et
al.~\cite{adebayo2018sanity} showed several saliency methods can be insensitive to
the model and the data they purport to explain, and argued for sanity checks
before any map is trusted. Rudin~\cite{rudin2019stop} argued more sharply that
high-stakes decisions call for inherently interpretable models rather than
post-hoc explanations of black boxes, and Doshi-Velez and
Kim~\cite{doshivelez2017rigorous} called for a measurable science of
interpretability. MIRROR is squarely the kind of system these critiques target,
and we take up its position against them in Section~\ref{sec:discussion} rather
than citing them as background.

\paragraph{Report generation.} Systems emitting radiology prose range from
TieNet~\cite{wang2018tienet}, which jointly embeds image and text, to
memory-driven transformers such as R2Gen~\cite{chen2020r2gen}, to decoders
optimized for factual correctness~\cite{liu2019clinically}. The recurring failure
mode is hallucination: prose generated directly from pixels can assert findings
the model never detected. Surveys of explainable AI in medical
imaging~\cite{tjoa2021survey} note that interpretability and report generation are
usually studied separately. MIRROR's contribution is neither a new classifier nor
a new saliency method but their composition under a constraint that bounds what
the report may assert.

\section{System Architecture}
\label{sec:method}

\begin{figure}[tbp]
\centering
\begin{tikzpicture}[
  node distance=5mm,
  layer/.style={rectangle, rounded corners=2.5pt, draw=mirrorblue!70, fill=boxbg,
                text width=0.82\linewidth, align=center, inner sep=4pt,
                font=\footnotesize},
  io/.style={rectangle, rounded corners=2pt, draw=mirrorgray, fill=white,
             text width=0.66\linewidth, align=center, inner sep=3pt,
             font=\footnotesize\itshape},
  arr/.style={-{Stealth[length=2.4mm]}, thick, mirrorgray}
]
\node[io] (img) {study + modality \\ \scriptsize(PNG/JPEG/BMP/WEBP/DICOM;
   modality explicit or from the DICOM header)};
\node[layer, below=of img] (clf)
  {\textbf{1. Classification}\\ \scriptsize CNN/ViT backbone, $N$-way multi-label
   head ($N$ sized to the modality) $\rightarrow$ per-label probabilities};
\node[layer, below=of clf] (loc)
  {\textbf{2. Evidence localization}\\ \scriptsize Grad-CAM/Score-CAM
   $\rightarrow$ heatmap + named region (lung zones or lobar regions)};
\node[layer, below=of loc] (rep)
  {\textbf{3. Clinical reasoning}\\ \scriptsize LLM or offline template over the
   \textbf{structured evidence only, never the pixels}};
\node[io, below=of rep] (out) {\textsc{Findings} / \textsc{Impression} report};
\draw[arr] (img) -- (clf);
\draw[arr] (clf) -- (loc);
\draw[arr] (loc) -- (rep);
\draw[arr] (rep) -- (out);
\end{tikzpicture}
\caption{The MIRROR pipeline as deployed in the local PyTorch stack, which is the
configuration benchmarked in Section~\ref{sec:results}. Layers 2 and 3 are
individually toggleable, which recovers the conditions of
Section~\ref{sec:setup}. The hosted engine of Section~\ref{sec:serving} does not
have this structure.}
\label{fig:arch}
\end{figure}
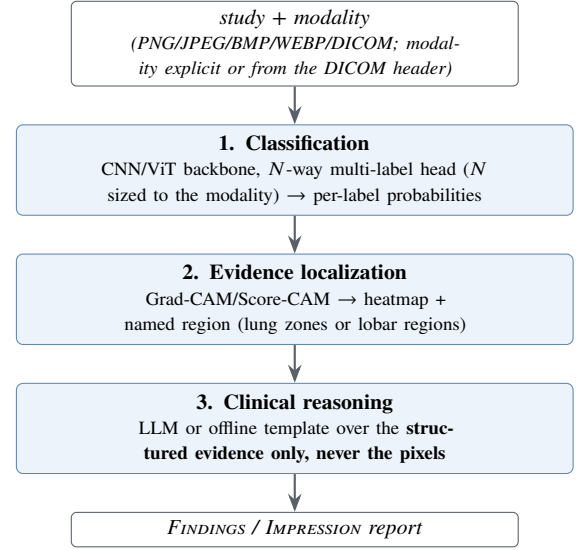

MIRROR chains three layers; each layer's output is the grounded input to the next
(Fig.~\ref{fig:arch}).

\subsection{Layer 1: Classification}
The classifier is a factory over three ImageNet-pretrained backbones: DenseNet-121
(default), EfficientNet-B0, and ViT-B/16, with a multi-label head whose width $N$
is the size of the active modality's taxonomy (14 for chest X-ray, 11 each for
brain MRI and head CT). The sigmoid is applied at loss and inference time rather
than inside the module, so the network emits raw logits. Inputs are resized to the
backbone's $224\times224$ input and normalized with ImageNet statistics; training uses
\code{BCEWithLogitsLoss}, AdamW, a cosine schedule, and dropout $0.2$, selecting
the checkpoint on best validation macro AUROC.

\subsection{DICOM ingest}
Real radiology data arrives as DICOM, and the gap
between that and a PNG is not a file-format detail. Stored pixel values are not
display values: they must pass through the modality LUT (the rescale slope and
intercept, which is what turns raw CT counts into Hounsfield units) and then the
VOI or window LUT that selects the diagnostic contrast range, and a
\code{MONOCHROME1} photometric interpretation means the stored values are inverted
relative to the usual convention. Skipping any of these silently trains a model on
images no radiologist would recognize, most visibly on \code{MONOCHROME1} studies,
which arrive as photographic negatives of what they should be. MIRROR ships an
ingest that applies all three transformations before the pipeline sees a pixel,
lifts only non-PHI technical tags, and exposes the modality tag for routing.

\subsection{Layer 2: Evidence Localization}
For each positive label (top-$k$, $k{=}3$ by default, to bound compute), this layer
computes a class-activation map, either Grad-CAM (forward and backward hooks on a
per-backbone target layer) or Score-CAM (gradient-free, perturbation-based); ViT
maps are recovered by reshaping the patch-token sequence back to a $14\times14$
spatial grid. The activated region's centroid is mapped into a $3\times3$
anatomical grid in radiology convention, with patient-right on the viewer's left,
and the modality's imaging plane picks the vocabulary: lung zones for a frontal
chest film, lobar regions for an axial brain-MRI or head-CT slice. A colormap
overlay is rendered for the interface, and the region name, not the heatmap, is
what travels downstream.

\subsection{Layer 3: Clinical Reasoning}
This layer converts structured evidence into a report with \textsc{Findings} and
\textsc{Impression} sections, using a system prompt plus an evidence-grounded user
prompt. Fig.~\ref{fig:evidence} shows the entire payload it receives: one line per
label, carrying the label itself, a plain-English gloss drawn from the registry
(Consolidation is glossed as ``region of lung filled with liquid instead of
air''), the probability, a present or below-threshold status, and the region name
from Layer 2. There is no image in the payload and no channel by which one could
arrive.

The backend is an LLM (Claude) at low temperature, with a deterministic offline
template fallback on any error, so the system yields a coherent report with no
network or API key. Predictions below the confidence threshold ($0.5$) surface as
pertinent negatives rather than being dropped, which keeps the uncertainty visible
instead of rounding it away. Because both backends consume identical evidence,
swapping them changes \emph{how fluently} the report reads, not which findings it
asserts. The grounding is therefore at the level of findings, not of the prose
framing them, a distinction that carries most of this paper's argument
(Section~\ref{sec:discussion}).

\begin{figure}[tbp]
\centering
\fbox{\begin{minipage}{0.92\linewidth}
\footnotesize\ttfamily
Modality: Chest X-ray\\
Clinical indication: <text, optional>\\
<per-modality report guidance>\\[3pt]
Structured model evidence (probabilities\\
from an image classifier, locations from\\
a Grad-CAM saliency map):\\
- <Label> (<gloss>): probability=<p>,\\
\hspace*{2ex}status=PRESENT, localised to <region>\\
- <Label> (<gloss>): probability=<p>,\\
\hspace*{2ex}status=below threshold, localised to n/a\\
\hspace*{2ex}... one line per label in the taxonomy
\end{minipage}}
\caption{The complete input to Layer 3. The reasoning layer sees this text and
nothing else: no pixels, no file handle, no image embedding. The grounding
property is a consequence of this payload's shape rather than of prompt wording,
which is why it can be asserted by a test instead of measured by a benchmark.}
\label{fig:evidence}
\end{figure}

\subsection{Multi-modality via a modality registry}
Swapping a chest radiograph for a brain MRI or head CT changes only the taxonomy
the head predicts, the vocabulary a saliency centroid is named in, and a little
report phrasing, so that a normal brain study reads ``no acute intracranial
abnormality'' rather than ``no acute cardiopulmonary abnormality.'' These live in
one registry (Table~\ref{tab:modalities}), the single source of truth for each
modality's labels, glosses, imaging plane, and report guidance; a hand-kept
TypeScript mirror gives the hosted engine and the web interface the identical
taxonomy and \emph{ordering}, since output index $i$ maps to
$\text{labels}[i]$. Modality is resolved from an explicit selection or, for DICOM,
from the (0008,0060) tag, and the orchestrator builds and caches one classifier
and explainer engine per modality on demand.

The claim here is architectural and is scoped as such. Routing, head sizing,
anatomical vocabulary, and per-modality phrasing are exercised by the test suite
for all three taxonomies, and the post-hoc invariance of Layers 2 and 3 is
asserted per modality. No brain-MRI or head-CT checkpoint has been trained, so
MIRROR makes no predictive claim on those modalities; what the registry
demonstrates is that adding a fourth would be a data change rather than a change
to the pipeline.

\begin{table}[tbp]
\centering
\caption{The modality registry. Registration means routing, head sizing, and
anatomical vocabulary are implemented and tested. Only chest X-ray is trained and
benchmarked here.}
\label{tab:modalities}
\footnotesize
\begin{tabular}{@{}p{0.20\linewidth}r%
>{\raggedright\arraybackslash}p{0.29\linewidth}%
>{\raggedright\arraybackslash}p{0.22\linewidth}@{}}
\toprule
\textbf{Modality} & \textbf{\#lbl} & \textbf{Taxonomy source} & \textbf{Location vocab.} \\
\midrule
Chest X-ray & 14 & NIH ChestX-ray14 & lung zones (frontal) \\
Brain MRI   & 11 & Brain Tumor MRI + neuro findings & lobar regions (axial) \\
Head CT     & 11 & RSNA ICH subtypes + acute findings & lobar regions (axial) \\
\bottomrule
\end{tabular}
\end{table}

\subsection{Orchestration and serving}
\label{sec:serving}
\begin{figure*}[tp]
\centering
\figorbox{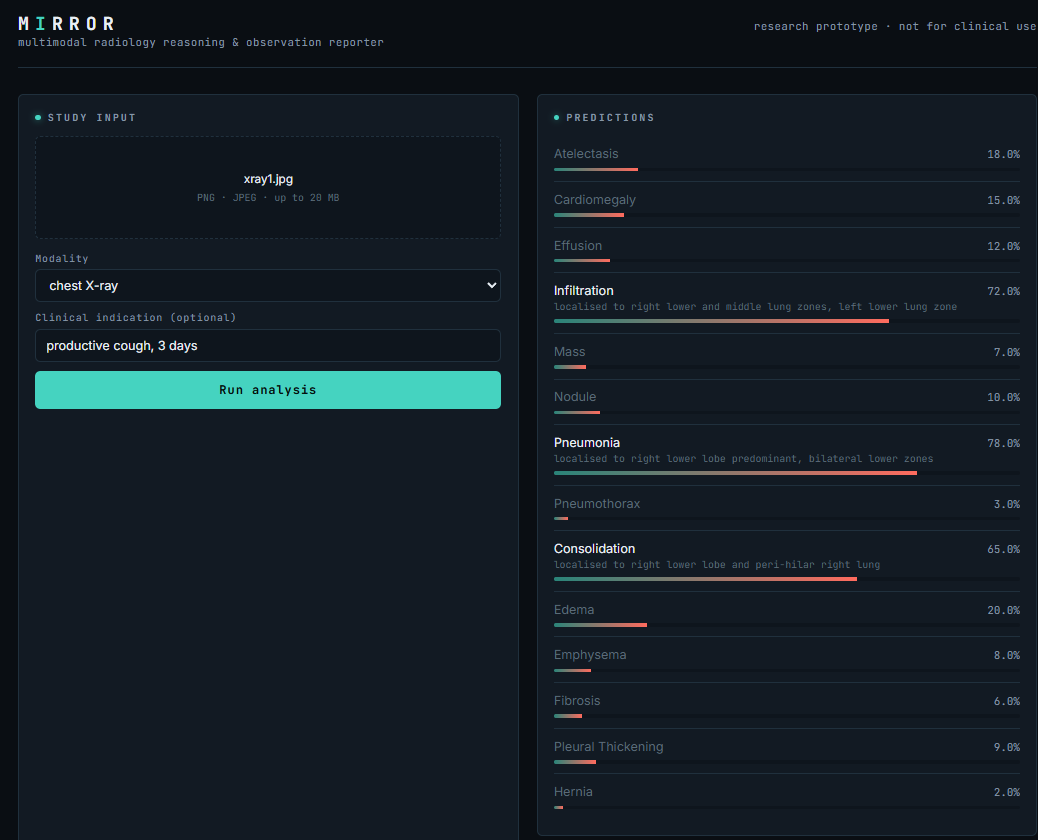}{0.56\textwidth}{MIRROR reading-room UI screenshot}
\caption{The MIRROR ``reading-room'' interface, showing a session on the
\textbf{hosted serverless engine}, where a vision LLM reads the image directly and
returns probabilities, boxes, and report text. This is \emph{not} the benchmarked
DenseNet-121 local stack, and no number in this figure is a measured result: it
illustrates the interface and the response contract only. A chest radiograph is
uploaded with the indication \emph{``productive cough, 3 days,''} all 14
ChestX-ray14 labels are scored, and findings above threshold surface with a
saliency-derived location. Both engines drive this identical interface.}
\label{fig:ui}
\end{figure*}

A single \code{analyze()} entry point runs the stages, records per-stage timings,
and returns one result object; Layers 2 and 3 are individually toggleable. Two
engines satisfy the same JSON contract (Table~\ref{tab:topology}), so the web
interface (Fig.~\ref{fig:ui}) is identical either way: a finding carries
\emph{either} a rendered Grad-CAM overlay from the local stack \emph{or} a
normalized bounding box from the hosted one, and the viewer draws whichever is
present.

The two engines are not two implementations of one design, and the difference
matters for every claim in this paper. The local stack is the architecture of
Fig.~\ref{fig:arch}: a trained DenseNet-121, a Grad-CAM map, and a language layer
that never receives pixels. The hosted engine exists because that stack does not
fit serverless limits, and it replaces the whole pipeline with a single vision LLM
under a forced tool call returning the per-label probabilities, one box per
present finding, and the report in one response. That engine reads pixels
directly, so MIRROR's grounding constraint \textbf{does not hold on the hosted
path}: nothing structurally prevents its prose from asserting a finding its own
probability vector does not support. Everything benchmarked in
Section~\ref{sec:results} is the local stack, and the two hosted-engine figures
are labelled as such.

\begin{table}[tbp]
\centering
\caption{Two inference engines behind one response contract. Only the local stack
implements the grounded architecture of Fig.~\ref{fig:arch}, and only it is
benchmarked here.}
\label{tab:topology}
\footnotesize
\begin{tabular}{@{}p{0.23\linewidth}%
>{\raggedright\arraybackslash}p{0.32\linewidth}%
>{\raggedright\arraybackslash}p{0.30\linewidth}@{}}
\toprule
 & \textbf{Local stack} & \textbf{Hosted (serverless)} \\
\midrule
Engine        & FastAPI + PyTorch      & Next.js route, vision LLM \\
Classify      & DenseNet/EffNet/ViT    & LLM scores $N$ labels \\
Localize      & Grad-CAM/Score-CAM PNG & LLM returns a bbox \\
Report        & LLM or template        & LLM, same call \\
Sees pixels?  & Layer 1 only           & Every stage \\
Inputs        & \small +DICOM, BMP     & PNG/JPEG/WEBP \\
Needs         & Python, $\sim$6\,GB    & one API-key variable \\
\bottomrule
\end{tabular}
\end{table}

\begin{figure*}[tp]
\centering
\begin{minipage}[t]{0.29\textwidth}
\centering
\figorbox{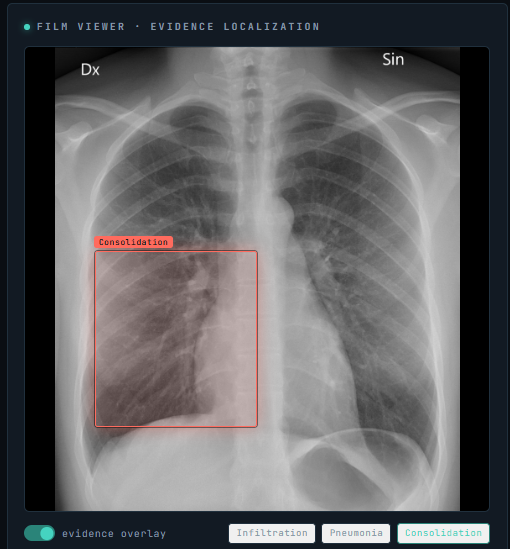}{\linewidth}{Grad-CAM evidence overlay}
\\[3pt]{\footnotesize (a) Evidence localization}
\end{minipage}\hspace{0.04\textwidth}%
\begin{minipage}[t]{0.29\textwidth}
\centering
\figorbox{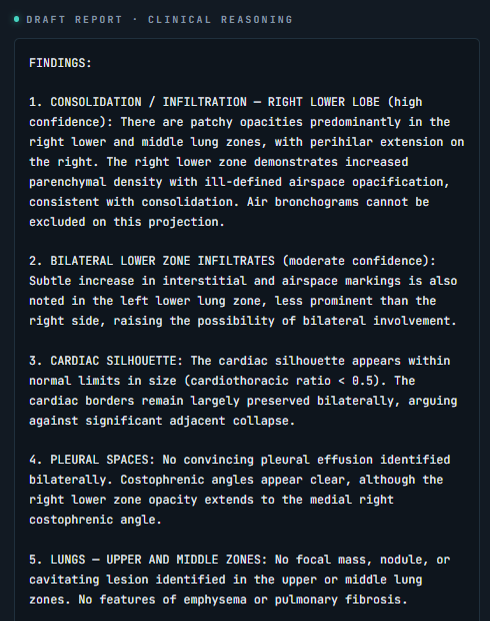}{\linewidth}{Draft report}
\\[3pt]{\footnotesize (b) Draft report}
\end{minipage}
\caption{One study end to end, also on the \textbf{hosted serverless engine} and
not the benchmarked local stack. (a) The localization layer marks the evidence
region behind a positive finding and the viewer draws it over the film. (b) The
reasoning layer drafts a \textsc{Findings}/\textsc{Impression} report ending with
the mandatory AI-generated-draft disclaimer. The report in (b) is the worked
example of Section~\ref{sec:ethics}: alongside its grounded findings it states a
cardiothoracic ratio and clear costophrenic angles, neither of which any component
of the system measured.}
\label{fig:qual}
\end{figure*}

\section{Experimental Setup}
\label{sec:setup}

\paragraph{Data.} MIRROR targets NIH ChestX-ray14~\cite{wang2017chestxray}
($112{,}120$ radiographs, 14 multi-label pathologies). Because the full release is
$45$\,GB and needs a GPU, we evaluate on \textbf{ChestMNIST}~\cite{yang2023medmnist},
its downsampled derivative in MedMNIST~v2 (CC~BY~4.0), which carries the
\emph{identical} 14-label taxonomy in the identical order, so MIRROR runs on it
unchanged. We sample $8{,}000$ studies from the pooled official train and
validation splits and hold out $10\%$, giving $7{,}200$ training and $800$
validation images, and evaluate on $12{,}000$ test images, a fixed random
subsample (seed $42$) of the official $22{,}433$-image test split, which is
otherwise held out untouched. Training is
DenseNet-121 on CPU, AdamW at learning rate $3\times10^{-4}$, weight decay
$10^{-5}$, batch $32$, best of a four-epoch run on validation macro AUROC. Source
images are $64\times64$ and are \emph{upsampled} to the backbone's $224\times224$
input: there is no genuine $224$-pixel information anywhere in this experiment,
and findings whose radiographic signature is a few millimetres wide are largely
destroyed before training begins.

\paragraph{Metrics and their floors.} We report a clinical-reader panel.
\emph{Discrimination:} per-label and macro AUROC, and macro AUPRC.
\emph{Operating point:} sensitivity, specificity, PPV, and NPV at the $0.5$
threshold, the quantities a clinician actually reads off a model.
\emph{Calibration:} Brier score and Expected Calibration
Error~\cite{guo2017calibration} over $10$ equal-width bins. Two of these are uninterpretable in isolation on an imbalanced
multi-label task, so we compute their no-skill floors from the label prevalences
and report both against them: a random ranker's expected average precision equals
the prevalence $p$, and a constant predictor that always returns $p$ scores a
Brier of exactly $p(1-p)$. Each headline metric carries a bootstrap $95\%$
confidence interval: the test set is resampled with replacement $B{=}1000$ times,
with a \emph{single} resample driving all metrics so the intervals are mutually
consistent, reported as two-sided percentile intervals ($\alpha{=}0.05$).

\paragraph{Post-hoc regression test.} Three conditions run the same studies with
Layers 2 and 3 switched on and off: classification only, classification with
localization, and the full pipeline. Those layers modify no weights and no logits,
so the predictions must be identical across conditions; the harness verifies this
and profiles the per-stage wall-clock cost of each added layer. This is a test
that the implementation matches the design, not an experiment with an uncertain
outcome, and Section~\ref{sec:ablation} reports it as such. Every results file
stamps a reproducibility record: seed, git commit, and library versions.

\section{Results}
\label{sec:results}

Every number below is measured on this machine with the local PyTorch stack and
traces to a JSON committed in the repository. Table~\ref{tab:panel} is the full
per-label panel and is the reference for Sections~\ref{sec:chestmnist}
through~\ref{sec:calib}.

\begin{table*}[tp]
\centering
\caption{\textbf{The full clinical-reader panel} on the ChestMNIST test split
(DenseNet-121, $64$\,px source upsampled to $224$\,px, CPU, best of four epochs;
$N_{\text{test}}{=}12{,}000$), sorted by AUROC. \emph{Prev.} is
$\text{support}_{+}/N$. \emph{Lift} is AUPRC divided by prevalence, the average
precision a uniformly random ranker would obtain, so it is the factor by which the
model beats chance at ranking that label. Every label ranks above chance; eleven of
them emit no positive prediction at the $0.5$ threshold, which is why their
sensitivity is $0.000$, their specificity $1.000$, and their PPV undefined. Hernia
rests on $24$ positive cases and is the least reliable row. Lifts use unrounded
AUPRC and prevalence, so a row's Lift can differ from the ratio of its displayed
values (Fibrosis $1.6$, Hernia $5.4$). The \textbf{Macro} Lift is the mean of the
per-label lifts, not macro AUPRC over macro prevalence ($2.6\times$); the Macro PPV
averages the eleven undefined labels as zero, and is $0.455$ over the three that
fire.}
\label{tab:panel}
\footnotesize
\begin{tabular}{@{}lcccccccc@{}}
\toprule
\textbf{Pathology} & \textbf{Prev.} & \textbf{AUROC} & \textbf{95\% CI} & \textbf{AUPRC} & \textbf{Lift} & \textbf{Sens.} & \textbf{Spec.} & \textbf{PPV} \\
\midrule
Edema              & 0.018 & 0.849 & (0.828--0.871) & 0.104 & 5.8$\times$ & 0.000 & 1.000 & --- \\
Cardiomegaly       & 0.027 & 0.830 & (0.806--0.851) & 0.184 & 6.8$\times$ & 0.000 & 1.000 & --- \\
Effusion           & 0.121 & 0.827 & (0.817--0.838) & 0.413 & 3.4$\times$ & 0.143 & 0.987 & 0.599 \\
Consolidation      & 0.041 & 0.754 & (0.736--0.773) & 0.092 & 2.2$\times$ & 0.000 & 1.000 & --- \\
Emphysema          & 0.023 & 0.744 & (0.713--0.774) & 0.069 & 3.0$\times$ & 0.000 & 1.000 & --- \\
Pneumothorax       & 0.049 & 0.738 & (0.719--0.757) & 0.138 & 2.8$\times$ & 0.012 & 0.999 & 0.350 \\
Atelectasis        & 0.109 & 0.729 & (0.716--0.742) & 0.221 & 2.0$\times$ & 0.000 & 1.000 & --- \\
Hernia             & 0.002 & 0.725 & (0.615--0.820) & 0.011 & 5.4$\times$ & 0.000 & 1.000 & --- \\
Mass               & 0.049 & 0.696 & (0.673--0.719) & 0.125 & 2.6$\times$ & 0.000 & 1.000 & --- \\
Fibrosis           & 0.015 & 0.674 & (0.639--0.705) & 0.025 & 1.6$\times$ & 0.000 & 1.000 & --- \\
Pneumonia          & 0.011 & 0.672 & (0.627--0.716) & 0.022 & 2.0$\times$ & 0.000 & 1.000 & --- \\
Pleural Thickening & 0.032 & 0.666 & (0.638--0.692) & 0.066 & 2.1$\times$ & 0.000 & 1.000 & --- \\
Infiltration       & 0.175 & 0.660 & (0.648--0.673) & 0.298 & 1.7$\times$ & 0.110 & 0.967 & 0.415 \\
Nodule             & 0.059 & 0.643 & (0.622--0.662) & 0.123 & 2.1$\times$ & 0.000 & 1.000 & --- \\
\midrule
\textbf{Macro} & & \textbf{0.729} & \textbf{(0.718--0.738)} & \textbf{0.135} & \textbf{3.1$\times$} & \textbf{0.019} & \textbf{0.997} & \textbf{0.097} \\
\multicolumn{9}{@{}l}{Macro F1 $0.031$ \qquad Brier $0.045$ against a constant-prevalence floor of $0.047$ \qquad ECE $0.018$ \qquad bootstrap $B{=}1000$} \\
\bottomrule
\end{tabular}
\end{table*}

\subsection{Discrimination on ChestMNIST}
\label{sec:chestmnist}
The DenseNet-121 reaches macro AUROC $0.729$ (bootstrap $95\%$ CI
$[0.718,0.738]$) on the $12{,}000$-image test split, with macro AUPRC $0.135$ and
macro F1 $0.031$. The published MedMNIST~v2 baseline for this dataset, a
ResNet-18/50 at $224$\,px (v2 reports no DenseNet-121), is approximately
$0.77$~\cite{yang2023medmnist}; ours is $0.04$ AUROC below it. That ResNet
baseline was trained on the full training split for many epochs on a GPU,
while ours used $7{,}200$ images for four epochs on a CPU. Both statements are
facts and we draw no inference from putting them side by side: a smaller budget
explains nothing about whether the gap would close, and we did not test that.

\begin{figure}[tbp]
\centering
\begin{tikzpicture}
\begin{axis}[
  width=0.88\linewidth, height=5.6cm,
  xbar, xmin=0.55, xmax=0.90,
  bar width=4.2pt,
  enlarge y limits=0.05,
  symbolic y coords={Nodule,Infiltration,Pleural Thickening,Pneumonia,Fibrosis,
    Mass,Hernia,Atelectasis,Pneumothorax,Emphysema,Consolidation,Effusion,
    Cardiomegaly,Edema},
  ytick=data,
  xlabel={AUROC (with bootstrap 95\% CI)},
  xmajorgrids, grid style={gray!25},
  tick label style={font=\scriptsize},
  label style={font=\footnotesize},
  axis line style={mirrorgray},
]
\addplot[fill=mirrorblue!35, draw=mirrorblue, error bars/.cd,
         x dir=both, x explicit] coordinates {
  (0.643,Nodule)             +- (0.020,0)
  (0.660,Infiltration)       +- (0.013,0)
  (0.666,Pleural Thickening) +- (0.027,0)
  (0.672,Pneumonia)          +- (0.044,0)
  (0.674,Fibrosis)           +- (0.033,0)
  (0.696,Mass)               +- (0.023,0)
  (0.725,Hernia)             +- (0.103,0)
  (0.729,Atelectasis)        +- (0.013,0)
  (0.738,Pneumothorax)       +- (0.019,0)
  (0.744,Emphysema)          +- (0.031,0)
  (0.754,Consolidation)      +- (0.019,0)
  (0.827,Effusion)           +- (0.011,0)
  (0.830,Cardiomegaly)       +- (0.023,0)
  (0.849,Edema)              +- (0.022,0)
};
\end{axis}
\end{tikzpicture}
\caption{Per-label test AUROC on ChestMNIST (DenseNet-121, $64$\,px source
upsampled to $224$\,px, CPU, best of four epochs, $N_{\text{test}}{=}12{,}000$),
with bootstrap $95\%$ CI whiskers, sorted ascending. The ordering tracks how
visible each finding is, and every interval clears the $0.5$ chance line. Eleven
of these labels nonetheless emit no positive prediction at the default threshold
(Section~\ref{sec:operating}).}
\label{fig:cxr-auroc}
\end{figure}
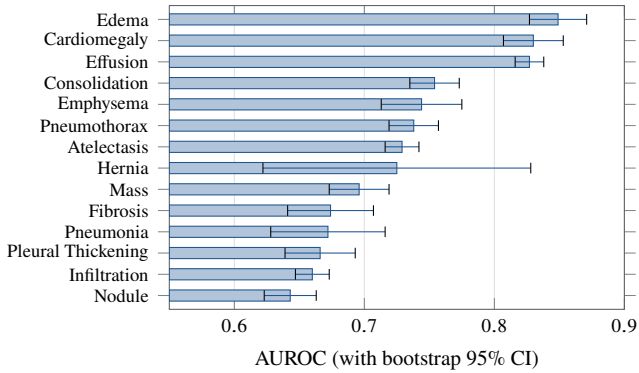

What the per-label ordering does support is that the model learned thoracic
structure rather than a dataset artifact (Fig.~\ref{fig:cxr-auroc}). Readily
visible findings rank highest, with Edema at $0.849$, Cardiomegaly at $0.830$, and
Effusion at $0.827$, while subtle small ones sit lowest, with Nodule at $0.643$
and Infiltration at $0.660$. Every confidence interval clears $0.5$. The widest
interval by a wide margin belongs to Hernia, $[0.615,0.820]$, which has only $24$
positive cases in the test split and should be read as the least reliable row in
the panel.

\subsection{The operating point: silence on most of the taxonomy}
\label{sec:operating}
AUROC is threshold-free, and it hides what this model does when asked for a
decision. At the default $0.5$ threshold, \textbf{11 of the 14 labels never
produce a single positive prediction} across all $12{,}000$ test studies: their
sensitivity is exactly $0.0$, their specificity exactly $1.0$, and their PPV
undefined. Only Effusion, Infiltration, and Pneumothorax fire at all, and
Pneumothorax fires on $1.2\%$ of its positive cases. Macro sensitivity is $0.019$
and macro F1 is $0.031$.

This is not a conservative operating point chosen for a clinical reason. Nothing
was tuned: $0.5$ is the default, and under prevalences running from $17.5\%$ for
Infiltration down to $0.2\%$ for Hernia, a lightly trained model minimizes its
loss by pushing almost every output below that line. At this threshold the
classifier is not usable as a detector. That is a property of the training budget
and the destroyed resolution, not a design decision.

\subsection{Ranking is real where decisions are absent}
\label{sec:auprc}
The natural next question is whether the eleven silent labels are silent because
the model learned nothing about them. They are not, and separating those two
explanations is the most useful thing the panel does. A uniformly random ranker's
expected average precision equals the label's
prevalence~\cite{davis2006pr,saito2015prc}, so AUPRC only means
something as a ratio to that floor. Expressed as lift over prevalence, every one
of the 14 labels ranks better than chance: from $1.6\times$ for Fibrosis to
$6.8\times$ for Cardiomegaly, with a mean of $3.1\times$
(Table~\ref{tab:panel}). Cardiomegaly is a clean illustration. Its prevalence is
$0.027$, its AUPRC is $0.184$, its AUROC is $0.830$ with a confidence interval
nowhere near chance, and it produces exactly zero positive predictions.

The model has therefore learned a usable ordering on every label in the taxonomy
and converts that ordering into a decision on almost none of them. The failure is
located at the threshold and in the calibration of the output scores, not in the
representation, which is a considerably more specific diagnosis than ``the model
underperforms'' and points at a different fix: per-label threshold selection on a
validation split, not more capacity.

\subsection{Calibration against a no-skill baseline}
\label{sec:calib}
Read on their own, the calibration numbers look strong: macro Brier $0.045$ and
ECE $0.018$. They are also close to meaningless here, and the arithmetic showing
it needs no new experiment. For a label with prevalence $p$, a constant predictor
that ignores the image and always returns $p$ achieves a Brier score of exactly
$p(1-p)$. Averaged over the 14 ChestMNIST prevalences, that is a macro Brier of
$\mathbf{0.0472}$ for a model that has learned nothing at all. Our trained
classifier scores $\mathbf{0.0453}$. The entire measured calibration advantage of
a DenseNet-121 over a constant is $0.0019$ Brier, about $4\%$.

That aggregate calibration metrics flatter a model under class imbalance is a
documented pitfall, not our finding; the arithmetic above makes it concrete.
Under the class
imbalance typical of multi-label radiology, where
most labels sit below $5\%$ prevalence, $p(1-p)$ is small for almost every label,
so the no-skill floor of the aggregate metric is already low and there is little
room between it and a perfect score. A model silent on 11 of 14 labels still posts
a Brier of $0.045$ and an ECE of $0.018$, numbers that would pass unremarked in
most results tables. Aggregate calibration metrics on imbalanced multi-label
medical tasks should be reported against the constant-prevalence baseline, or not
offered as evidence of quality at all.

\subsection{Harness control on synthetic data}
\label{sec:synthetic}
Numbers are only worth reporting if the harness that produced them credits real
signal and nothing else, so we test that directly. We train the same pipeline on a
fully synthetic dataset of $1{,}400$ procedurally generated chest-radiograph-like
images whose generator injects a visible blob for seven named labels, Mass,
Nodule, Effusion, Infiltration, Consolidation, Edema, and Cardiomegaly, and leaves
the other seven with no visual correlate whatsoever. The seven are fixed in the
generator in advance, not chosen after seeing results.

\begin{figure}[tbp]
\centering
\begin{tikzpicture}
\begin{axis}[
  width=0.88\linewidth, height=5.6cm,
  xbar, xmin=0.35, xmax=1.02,
  bar width=4.2pt,
  enlarge y limits=0.05,
  symbolic y coords={Pleural Thickening,Pneumothorax,Atelectasis,Pneumonia,
    Emphysema,Fibrosis,Hernia,Edema,Infiltration,Effusion,Consolidation,
    Cardiomegaly,Nodule,Mass},
  ytick={Pleural Thickening,Pneumothorax,Atelectasis,Pneumonia,
    Emphysema,Fibrosis,Hernia,Edema,Infiltration,Effusion,Consolidation,
    Cardiomegaly,Nodule,Mass},
  xlabel={AUROC (synthetic test set, $N{=}350$)},
  xmajorgrids, grid style={gray!25},
  tick label style={font=\scriptsize},
  label style={font=\footnotesize},
  legend style={font=\scriptsize, at={(0.98,0.03)}, anchor=south east, draw=gray!40},
  axis line style={mirrorgray},
]
\addplot[fill=mirrorgray!35, draw=mirrorgray] coordinates {
  (0.434,Pleural Thickening) (0.467,Pneumothorax) (0.524,Atelectasis)
  (0.540,Pneumonia) (0.550,Emphysema) (0.595,Fibrosis) (0.620,Hernia)};
\addplot[fill=mirrorblue!55, draw=mirrorblue] coordinates {
  (0.866,Edema) (0.894,Infiltration) (0.895,Effusion) (0.903,Consolidation)
  (0.908,Cardiomegaly) (0.974,Nodule) (0.979,Mass)};
\legend{No injected signal, Injected visual signal}
\end{axis}
\end{tikzpicture}
\caption{Harness control on synthetic data. The generator injects a blob for seven
labels named in advance and none for the other seven; the classifier learns
\emph{exactly} the signal-bearing group (mean AUROC $0.917$) and stays near chance
on the rest (mean $0.533$), with no overlap between the groups. This checks that
the metrics measure real discrimination rather than label frequency, and that
nothing is leaking.}
\label{fig:synth}
\end{figure}
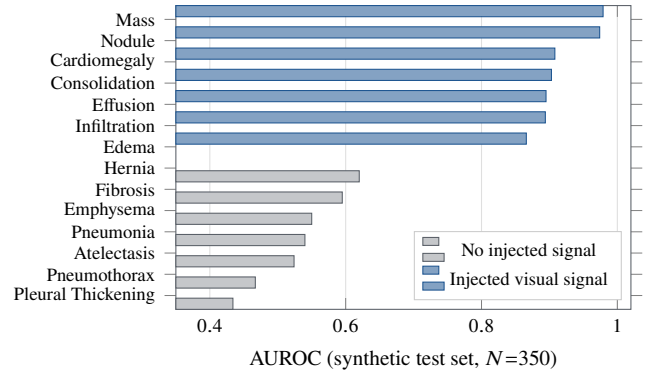

A correct harness should learn exactly the first group and sit near chance on the
second, and that is what happens (Fig.~\ref{fig:synth}). On a $350$-image held-out
synthetic test split the seven signal-bearing labels reach a mean AUROC of $0.917$,
from Mass at $0.979$ down to Edema at $0.866$, and the seven no-signal labels
average $0.533$, from Hernia at $0.620$ down to Pleural Thickening at $0.434$. The
two groups do not overlap: the worst signal-bearing label outscores the best
no-signal label by $0.25$ AUROC. The no-signal mean sits slightly above $0.5$
rather than exactly on it, which is what a test split this size should give, and
those labels straddle chance in both directions rather than clustering above it,
which is what rules out a systematic leak. The classifier, the loss, the metrics,
and the bootstrap intervals therefore respond to injected signal and not to label
frequency or ordering.

\subsection{Post-hoc invariance and latency}
\label{sec:ablation}
Layers 2 and 3 run after the forward pass and modify no weights and no logits, so
the predictions are identical across the three conditions by construction. The
harness confirms it: the maximum per-label probability change against the
classification-only baseline is $0.000$ over $n{=}24$ studies. This is a
regression test, not a result. It earns its place because a nonzero delta would
expose a real bug, an accidental coupling between the layers, and the test suite
asserts the same property per modality against stubbed backbones so it holds
structurally rather than only for the chest checkpoint.

The cost that does vary is latency (Table~\ref{tab:ablation}). Prediction takes
about $100$\,ms per study on CPU, the Grad-CAM pass adds $36$ to $41$\,ms, and the
offline template report adds $0.03$\,ms. The two conditions that run Grad-CAM
measured $41.4$ and $36.2$\,ms for the same work, which is the run-to-run spread of
a 24-study CPU profile and is why the full-pipeline total of $136.2$\,ms came in
below the localization-only total of $140.9$\,ms. Interpretability here costs
roughly a $40\%$ increase in wall-clock time over a bare classifier: all of that
increase sits in the saliency pass, and none of it in the report.

\begin{table}[tbp]
\centering
\caption{Ablation conditions with measured per-stage wall-clock latency
(ms/study, mean over $24$ studies on CPU). Each row is a separate profiling run,
so its \emph{Predict} time carries the run-to-run spread of a 24-study CPU
profile; within each row \emph{Predict $+$ Localize $+$ Report} equals
\emph{Total}. That spread, not any change in work, is why full MIRROR totals less
than localization alone. \textbf{There is one AUROC measurement here, not three:}
macro AUROC $=0.729$ from Table~\ref{tab:panel} applies unchanged to every row
because the post-hoc layers cannot alter a prediction.}
\label{tab:ablation}
\footnotesize
\begin{tabular}{@{}lrrrr@{}}
\toprule
\textbf{Condition} & \textbf{Predict} & \textbf{Localize} & \textbf{Report} & \textbf{Total} \\
\midrule
Classification only       & 101.2 & ---  & ---  & 101.2 \\
$+$ Evidence localization & \phantom{0}99.5 & 41.4 & ---  & 140.9 \\
Full MIRROR               & 100.0 & 36.2 & 0.03 & 136.2 \\
\midrule
\multicolumn{5}{@{}l}{\footnotesize Macro AUROC $=0.729$ in every row: one measurement.} \\
\multicolumn{5}{@{}l}{\footnotesize Max probability change vs.\ baseline $=0.000$, $n{=}24$.} \\
\bottomrule
\end{tabular}
\end{table}

\section{Discussion}
\label{sec:discussion}

\paragraph{Finding-level grounding is a real guarantee and a narrow one.} The
useful thing MIRROR demonstrates is a boundary, and stating it precisely is worth
more than the classifier behind it. Because the language layer receives a label, a
probability, and a region name rather than an image (Fig.~\ref{fig:evidence}), the
set of findings the report may assert is exactly the set the classifier produced.
That is checkable without trusting anything about the model: a reader holding the
report and the probability vector can verify the correspondence, and a unit test
can assert it. It is also strictly a claim about \emph{which findings appear}, and
says nothing about the sentences around them. A report that correctly asserts
consolidation may frame it with clauses the system never computed, and those
clauses inherit the credibility of the finding they sit beside
(Section~\ref{sec:ethics}). This distinction is the thing worth taking from the
system, and architectures that constrain only the former should say so rather than
claiming groundedness in general.

\paragraph{Where this stands against Rudin.} MIRROR is precisely the architecture
Rudin~\cite{rudin2019stop} argues against for high-stakes decisions: an opaque CNN
with a saliency map and a language layer attached after the fact. We have no
rebuttal, and we do not claim the saliency map is faithful; the sanity checks of
Adebayo et al.~\cite{adebayo2018sanity} are the right instrument for testing that
and we have not run them. What we claim is narrower and does not depend on the
heatmap being faithful at all: the finding set is auditable, because a reader can
check the report against the probability vector directly. That is a retreat from
explanation to auditability, and we would rather name it than dress it up. An
inherently interpretable classifier would be the stronger answer.

\paragraph{Discrimination and decisions are different failures.} The sequence of
numbers this model produces is instructive, and it is the reason we report the
panel in full. Macro AUROC $0.729$ reads as a modest but working model. Brier
$0.045$ and ECE $0.018$ read as a well-calibrated one. AUPRC lift of $1.6$ to
$6.8$ times prevalence says the ranking is genuinely informative on every single
label. And the model emits no positive prediction for 11 of 14 labels. All four
statements are true simultaneously, which they can be only because ranking quality
and decision quality are separate properties that the usual headline metrics do
not distinguish. Reporting AUROC alone would have hidden the failure; reporting
the operating point alone would have hidden that the representation is sound and
the fix is a per-label threshold rather than a bigger network. The practical
recommendation is to publish sensitivity and PPV alongside every AUROC in this
literature, and to write both AUPRC and Brier against their prevalence-derived
floors, which costs one line of arithmetic and changes how the numbers read.

\paragraph{One contract, two very different engines.} Serving the same JSON
response from a PyTorch stack and from a hosted vision LLM shows the
\emph{interface} MIRROR proposes, namely predict, show evidence, explain, is
independent of the inference backend, at the price of the asymmetry in
Section~\ref{sec:serving}. We would rather flag that than let one system's
architecture diagram vouch for the other's behaviour.

\section{Ethics, Safety, and Limitations}
\label{sec:ethics}

\paragraph{Not a medical device.} MIRROR is a research prototype, not reviewed or
cleared by any regulatory body; every output is a draft that must be verified by a
licensed radiologist and must not be used for diagnosis or treatment. Each report
carries that disclaimer explicitly.

\paragraph{A worked example of ungrounded detail.} Fig.~\ref{fig:qual} shows a
generated report beside its evidence overlay. Two phrases in it are worth quoting:
the cardiac silhouette is described as normal \emph{with a cardiothoracic ratio
below $0.5$}, and the \emph{costophrenic angles are noted to be clear}. The system
measured neither. No cardiothoracic ratio was computed, no cardiac or thoracic
boundary was segmented, and no costophrenic angle was localized; the pipeline
produced a probability per label and up to three saliency centroids, and nothing
in that evidence supports either clause. Both are fluent radiology register
generated to fill out a report. They also sit in the same paragraph as assertions
that \emph{are} grounded, and that adjacency is the danger: a reader who checks
the consolidation finding against the overlay and finds it correct has been given
a reason to extend trust to the sentence beside it. Fluency makes ungrounded
detail more dangerous rather than less, because it arrives wearing the credibility
of the verified finding next to it. This is why finding-level grounding cannot be
described as making a report safe, and why the AI-generated-draft disclaimer is
load-bearing rather than boilerplate.

\paragraph{Human subjects and data governance.} No ethics-board review was required:
this work uses only existing, public, de-identified benchmark datasets, namely NIH
ChestX-ray14 and its MedMNIST derivative, and collects no new data from human
subjects. Those datasets are governed by their own data use agreements; no raw
images, weights, or protected health information are committed to version control,
and the DICOM ingest extracts only non-PHI technical tags.

\paragraph{Limitations.} Our one quantitative benchmark is ChestMNIST, at $64$-pixel
source resolution on a $7{,}200$-image budget, and at its default threshold the
classifier is not a working detector (Section~\ref{sec:operating}). No clinical
claim of any kind follows from it. Multi-modality is architectural: the brain-MRI
and head-CT paths are implemented and exercised by the test suite but have no
trained checkpoints, so MIRROR makes no predictive claim on those modalities.
Explanation quality is unmeasured. The pointing-game and IoU protocol is
implemented but not run, and no reader study has been conducted, so we have no
evidence that the localization or report layers help a human, only that they cost
about $40$\,ms. Class-activation maps are coarse and can highlight context rather
than lesion~\cite{adebayo2018sanity}, and we have not run the sanity checks that
would detect it. The $3\times3$ location grid is a deliberately coarse anatomical
descriptor, and Hernia aside, the panel's narrowest intervals still rest on a
single training run rather than a multi-seed aggregate. Report quality is assessed
only qualitatively; a comparison against real radiologist reports, for example
MIMIC-CXR~\cite{johnson2019mimiccxr}, is future work.

\section{Conclusion}

MIRROR composes classification, evidence localization, and report generation into
one traceable, registry-driven pipeline, and the property worth carrying forward
is a boundary rather than a score: constraining the language layer to structured
evidence makes the \emph{finding set} of a report auditable and does nothing for
the prose around it. On ChestMNIST the classifier reaches macro AUROC $0.729$,
ranks every label between $1.6$ and $6.8$ times better than a random ranker, and
still emits no positive prediction for 11 of its 14 labels while posting a Brier
score within $0.002$ of a predictor that never looks at the image. We report those
together deliberately: they are the clearest evidence we produced that aggregate
metrics on imbalanced multi-label tasks can look healthy over a model that makes
no decisions. Next steps are full-resolution training on the NIH release,
per-label threshold selection on a validation split, the saliency sanity checks we
cite but did not run, and a reader study, which is the only thing that would test
whether any of this helps a radiologist.

\section*{Reproducibility}
\footnotesize
\begin{sloppypar}\raggedright
Code: \url{https://github.com/vignesh-nagarajan-vn/MIRROR}. Live demo, on the
hosted engine: \url{https://mirror-ten-jet.vercel.app/}. A zero-setup CLI demo
(\code{python -m demo.run\_demo <image>}) runs the local pipeline on ImageNet
weights with the offline template backend, needing no checkpoint and no API key.
Every results JSON stamps a \code{reproducibility} block with the seed, git
commit, and library versions, and the unit tests are torch-free. The ChestMNIST
result (\code{configs/chestmnist.yaml}) and the synthetic control
(\code{configs/synthetic.yaml}) regenerate from seed $42$; their outputs are
committed under \code{results/}. The prevalences, the AUPRC lifts of
Section~\ref{sec:auprc}, the no-skill Brier floor of Section~\ref{sec:calib}, and
the exact contents of Table~\ref{tab:panel} are recomputed from those committed
files by \code{paper/calibration\_baseline.py}.
\end{sloppypar}

\section*{Acknowledgments}
\small
This work was completed as part of the Global Indian Scientists \& Technocrats
(GIST) 2026 Summer Research Internship. The author thanks his mentor,
\textbf{Sriram Venkatapathy}, an Applied AI Researcher at Capital One who holds a
PhD in Computer Science from the Indian Institute of Technology (IIT) Hyderabad,
for his guidance throughout the project: his experience building and evaluating
machine-learning systems in industry shaped both the architecture and the
emphasis on measurable, reproducible evaluation.

\section*{Declarations}
\footnotesize
\textbf{Data availability:} all datasets used are public, namely NIH ChestX-ray14
and its MedMNIST derivative ChestMNIST (CC~BY~4.0); code, configurations, and result
files are in the repository linked above. \textbf{Competing interests:} the author
declares none. \textbf{Funding:} this work received no external research funding and
was conducted as part of the GIST 2026 Summer Research Internship.

\balance


\begin{thebibliography}{99}
\scriptsize
\setlength{\itemsep}{0pt}
\setlength{\parsep}{0pt}

\bibitem{wang2017chestxray}
X.~Wang, Y.~Peng, L.~Lu, Z.~Lu, M.~Bagheri, and R.~M. Summers.
ChestX-ray8: Hospital-scale chest X-ray database and benchmarks on
weakly-supervised classification and localization of common thorax diseases.
\emph{CVPR}, 2097--2106, 2017.

\bibitem{rajpurkar2017chexnet}
P.~Rajpurkar, J.~Irvin, K.~Zhu, et~al.
CheXNet: Radiologist-level pneumonia detection on chest X-rays with deep learning.
\emph{arXiv:1711.05225}, 2017.

\bibitem{irvin2019chexpert}
J.~Irvin, P.~Rajpurkar, M.~Ko, et~al.
CheXpert: A large chest radiograph dataset with uncertainty labels and expert
comparison. \emph{AAAI}, 2019.

\bibitem{johnson2019mimiccxr}
A.~E.~W. Johnson, T.~J. Pollard, S.~J. Berkowitz, et~al.
MIMIC-CXR, a de-identified publicly available database of chest radiographs with
free-text reports. \emph{Scientific Data}, 6(1):317, 2019.

\bibitem{yang2023medmnist}
J.~Yang, R.~Shi, D.~Wei, Z.~Liu, L.~Zhao, B.~Ke, H.~Pfister, and B.~Ni.
MedMNIST v2: A large-scale lightweight benchmark for 2D and 3D biomedical image
classification. \emph{Scientific Data}, 10(1):41, 2023.

\bibitem{huang2017densenet}
G.~Huang, Z.~Liu, L.~van~der~Maaten, and K.~Q. Weinberger.
Densely connected convolutional networks. \emph{CVPR}, 4700--4708, 2017.

\bibitem{tan2019efficientnet}
M.~Tan and Q.~V. Le.
EfficientNet: Rethinking model scaling for convolutional neural networks.
\emph{ICML}, 2019.

\bibitem{dosovitskiy2021vit}
A.~Dosovitskiy, L.~Beyer, A.~Kolesnikov, et~al.
An image is worth 16x16 words: Transformers for image recognition at scale.
\emph{ICLR}, 2021.

\bibitem{zhou2016cam}
B.~Zhou, A.~Khosla, A.~Lapedriza, A.~Oliva, and A.~Torralba.
Learning deep features for discriminative localization. \emph{CVPR},
2921--2929, 2016.

\bibitem{selvaraju2017gradcam}
R.~R. Selvaraju, M.~Cogswell, A.~Das, R.~Vedantam, D.~Parikh, and D.~Batra.
Grad-CAM: Visual explanations from deep networks via gradient-based localization.
\emph{ICCV}, 618--626, 2017.

\bibitem{wang2020scorecam}
H.~Wang, Z.~Wang, M.~Du, et~al.
Score-CAM: Score-weighted visual explanations for convolutional neural networks.
\emph{CVPR Workshops}, 2020.

\bibitem{ribeiro2016lime}
M.~T. Ribeiro, S.~Singh, and C.~Guestrin.
``Why should I trust you?'': Explaining the predictions of any classifier.
\emph{KDD}, 1135--1144, 2016.

\bibitem{lundberg2017shap}
S.~M. Lundberg and S.-I. Lee.
A unified approach to interpreting model predictions. \emph{NeurIPS}, 2017.

\bibitem{zhang2016pointing}
J.~Zhang, S.~A. Bargal, Z.~Lin, J.~Brandt, X.~Shen, and S.~Sclaroff.
Top-down neural attention by excitation backprop. \emph{ECCV}, 2016.

\bibitem{adebayo2018sanity}
J.~Adebayo, J.~Gilmer, M.~Muelly, I.~Goodfellow, M.~Hardt, and B.~Kim.
Sanity checks for saliency maps. \emph{NeurIPS}, 2018.

\bibitem{rudin2019stop}
C.~Rudin.
Stop explaining black box machine learning models for high stakes decisions and
use interpretable models instead. \emph{Nature Machine Intelligence},
1(5):206--215, 2019.

\bibitem{doshivelez2017rigorous}
F.~Doshi-Velez and B.~Kim.
Towards a rigorous science of interpretable machine learning.
\emph{arXiv:1702.08608}, 2017.

\bibitem{wang2018tienet}
X.~Wang, Y.~Peng, L.~Lu, Z.~Lu, and R.~M. Summers.
TieNet: Text-image embedding network for common thorax disease classification and
reporting in chest X-rays. \emph{CVPR}, 2018.

\bibitem{chen2020r2gen}
Z.~Chen, Y.~Song, T.-H. Chang, and X.~Wan.
Generating radiology reports via memory-driven transformer. \emph{EMNLP}, 2020.

\bibitem{liu2019clinically}
G.~Liu, T.-M.~H. Hsu, M.~McDermott, et~al.
Clinically accurate chest X-ray report generation. \emph{MLHC}, 2019.

\bibitem{tjoa2021survey}
E.~Tjoa and C.~Guan.
A survey on explainable artificial intelligence (XAI): Toward medical XAI.
\emph{IEEE TNNLS}, 32(11):4793--4813, 2021.

\bibitem{davis2006pr}
J.~Davis and M.~Goadrich.
The relationship between precision-recall and ROC curves. \emph{ICML},
233--240, 2006.

\bibitem{saito2015prc}
T.~Saito and M.~Rehmsmeier.
The precision-recall plot is more informative than the ROC plot when evaluating
binary classifiers on imbalanced datasets. \emph{PLOS ONE}, 10(3):e0118432, 2015.

\bibitem{guo2017calibration}
C.~Guo, G.~Pleiss, Y.~Sun, and K.~Q. Weinberger.
On calibration of modern neural networks. \emph{ICML}, 1321--1330, 2017.

\end{thebibliography}
\end{document}